\documentclass[runningheads]{llncs}
\usepackage{graphicx}
\usepackage{amsmath,amssymb} 
\usepackage{color}
\usepackage{wrapfig}
\graphicspath{{./images/}}

\begin{document}

\newcommand{\point}{
    \raise0.7ex\hbox{.}
    }

\pagestyle{headings}

\mainmatter

\title{Combining Texture and Shape Cues for Object Recognition With Minimal Supervision}

\titlerunning{Combining Texture and Shape Cues for Object Recognition}

\authorrunning{Xingchao Peng and Kate Saenko } 

\author{Xingchao Peng and Kate Saenko }
\institute{Computer Science Department, Boston University\\ \{xpeng, saenko\}@bu.edu}

\maketitle

\begin{abstract}
We present a novel approach to object classification and detection which requires minimal supervision and which combines visual texture cues and shape information learned from freely available unlabeled web search results. The explosion of visual data on the web can potentially make visual examples of almost any object easily accessible via web search. Previous unsupervised methods have utilized either large scale sources of texture cues from the web, or shape information from data such as crowdsourced CAD models. We propose a two-stream deep learning framework that combines these cues, with one stream learning visual texture cues from image search data, and the other stream learning rich shape information from 3D CAD models. To perform classification or detection for a novel image, the predictions of the two streams are combined using a late fusion scheme.  We present experiments and visualizations for both tasks on the standard benchmark PASCAL VOC 2007 to demonstrate that texture and shape provide complementary information in our model. Our method outperforms previous web image based models, 3D CAD model based approaches, and weakly supervised models.
\end{abstract}

\section{Introduction}
\label{intro}
Object classification and detection are fundamental tasks in computer vision. Previous mainstream object detectors based on hand-designed features (HOG\cite{dalal2005histograms}) and classifiers like linear discriminative analysis (LDA) \cite{welling2005fisher} or Support Vector Machines~\cite{boser1992training,lsvm-pami} required training of limited parameters, and were thus sustainable with small datasets. More recent Deep Convolutional Neural Network (DCNN)-based object detectors\cite{RCNN,renNIPS15fasterrcnn} utilize more powerful DCNN features and yield a significant performance boost, both for image classification\cite{he2015deep,vgg,alexnet,GoogleLenet} and object detection\cite{RCNN,renNIPS15fasterrcnn}. Nevertheless, deep convolutional neural networks need lots of labeled images to train their millions of parameters. Collecting these images and annotating the objects is cumbersome and expensive. Even the most popular detection datasets provide a limited number of labeled categories, e.g., 20 classes in PASCAL VOC\cite{pascal} and 200 in ImageNet\cite{ImageNet}. Hence, a question arises: is it possible to avoid the frustrating collection and annotation process and still train an effective object classifier or detector?

To achieve this goal, researchers have proposed several recent models for object model training. \cite{LSDA} introduces a transfer learning method that gains an object detector by transferring learned object knowledge from a classifier. \cite{siva2011weakly,song2014learning} propose to train an object localization and detection model with image-level labels. However, these methods still count on per-image manual supervision. In contrast, the methods in the previous literature that assume no per-image labeling data can be categorized into two groups: 1) methods that utilize on-line search results or an existing unlabeled dataset \cite{chen2015webly,divvala2014learning,zhou2015conceptlearner}; 2) methods that render domain-specific synthetic images \cite{peng2015learning,BMVC}. For example, \cite{chen2015webly,divvala2014learning} propose to learn a visual representation and object detectors from on-line images. \cite{zhou2015conceptlearner} leverages a concept learner to discover visual knowledge from weakly labeled images (weak labels can be in the form of keywords or short description). On the other hand, \cite{peng2015learning,BMVC} proposed to generate synthetic training images from 3D CAD models. Such synthetic data is shown to be useful for augmenting small amounts of real labeled data to improve object detection, as bounding boxes can be obtained ``for free'' and  objects can be rendered in arbitrarily many viewpoints. 

\begin{figure}[t]
    \centering
    \includegraphics[width=\textwidth]{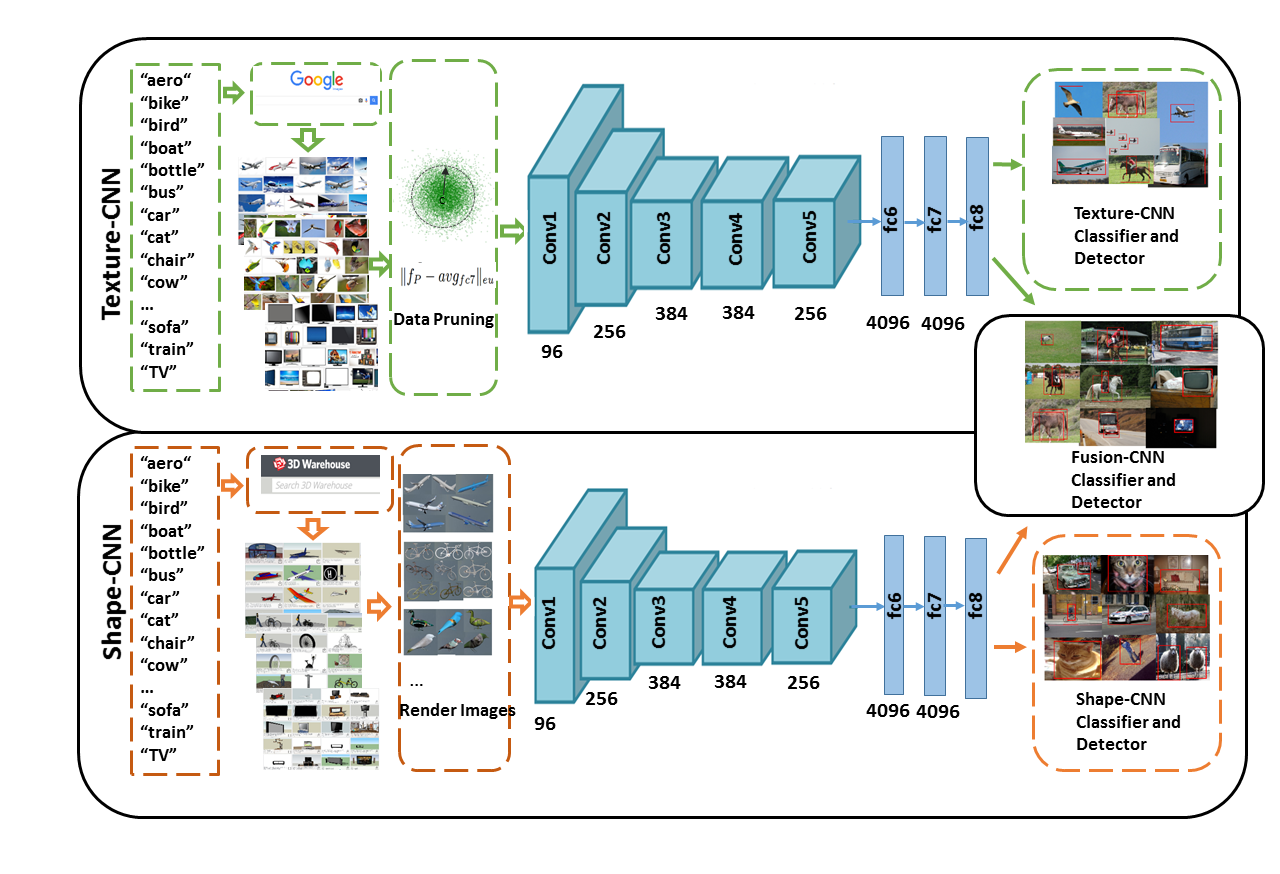}
    \vspace{-1cm}
    \caption{\textbf{Two-Stream Texture and Shape Model:}
    We propose a framework to combine texture cues and shape information for object recognition with minimal supervision. The Texture-CNN stream is trained on images collected from on-line image search engines and the Shape-CNN stream is separately trained on images generated from domain-specific 3D CAD models. To combine the two streams, an average of the two last layers' activations is computed to create the Fusion-CNN object classifier and detector. In the test phrase, the model will forward an image patch through the two networks simultaneously and compute average fusion of the activations from the last layers.  (Best viewed in color.) }
    \label{fig:overview}
    \vspace{-0.73cm}
\end{figure}

While these approaches can work effectively in some cases, there are still many challenges that need to be addressed:
\begin{itemize}
\item \textbf{Lack of bounding boxes:} Unsupervised machine learning algorithms tackle learning  where no labeled training data is provided. This setting leads to a great challenge for object detection because the performance of a detection system depends heavily on differentiated positive and negative training examples labeled with tight bounding boxes. Without such annotations, it is difficult for a model to learn the extent and shape of objects, i.e. which parts of the image correspond to the target object and which are background. 
\item \textbf{Missing shape or texture cues:} Prior literature uses either web search~\cite{chen2015webly,divvala2014learning} or synthetic images~\cite{peng2015learning,su2015render,BMVC}, but rarely combines the intrinsic cues like object shape and characteristic appearance patterns, or ``texture'', which are critical for recognition systems. Some rigid objects can be easily recognized by its shape, such as aeroplanes and sofa; other objects can be easily recognized by their unique texture, such as leopards and bees. 

\item \textbf{Domain Shift:} Images from different sources have different statistics for background, texture, intensity and even illumination \cite{saenko2010adapting}, which consequently results in domain shift problems. Unlike images taken in the wild, most photographs returned by image search engines lack diversity in viewpoint, background and shape, for image search engines follow a high-precision low-recall regime. In addition, synthetic images used in current work~\cite{peng2015learning} are far from photorealism in terms of intensity contrast, background, and object texture.  

\end{itemize}

In this paper, we address these shortcomings by proposing a two-stream DCNN architecture (See Figure \ref{fig:overview}) that decomposes the input into texture and shape feature streams. The texture stream learns realistic texture cues from images downloaded from the Internet. Web images (which contain noise from backgrounds and unrelated results) are collected by searching for the names of categories in image search engines, i.e. Google Image Search. We prune the noise data and use the cleaned data to train the texture-based stream for classification or detection. The shape stream is trained exclusively on shape information from 2D images rendered from CAD models annotated with category labels. 
Note that images from web search also contain shape information, but due to the lack of bounding box annotations, it is not accurate enough for localization models. Synthetic images can also be generated from 3D CAD models with added texture mapping, but the result is non-photorealistic and lacks real-world variety. Therefore, synthetic images rendered by 3D CAD models can be viewed as primarily shape-oriented and the web-search images can be viewed as primarily texture-oriented. The outputs of the two streams are combined through averaging the two top layers' activations. Our method requires no tedious manual bounding box annotation of object instances and no per-image category labeling and can generate training data for almost any novel category. Table \ref{tab_compare} shows a comparison of the amount of supervision with other methods. The only supervision in our work comes from labeling the CAD models as positive examples vs. ``outliers'' while downloading them, and choosing proper textures for each category while generating the synthetic images.

We evaluate our model for object classification and detection on the standard PASCAL VOC 2007 dataset, and show that the texture and shape cues can reciprocally compensate for each other during recognition. In addition, our detector outperforms existing webly supervised detectors~\cite{divvala2014learning}, the approach based on synthetic images only~\cite{peng2015learning}, and the weakly supervised learning approach of~\cite{siva2011weakly}, despite using less manual supervision.

\begin{table*}[t]

\scriptsize

\noindent\begin{tabular}{ |c | c|c|   }
\hline
Type &VOC Training Data Used&Supervision\\
\hline
CAD Supervision, Ours, \cite{peng2015learning} & NO & Labeling 3D CAD models and their texture\\
\hline
Webly Supervision \cite{chen2015webly,divvala2014learning}& NO & Semantic Labeling \\
\hline
Selected Supervision \cite{zhou2015conceptlearner}& NO & Per-image Labeling With Text \\
\hline
Weak Supervision \cite{siva2011weakly,song2014learning}& YES & Per-image Labeling \\
\hline
Full Supervision \cite{RCNN}& YES & Per-instance Labeling\\
\hline

\end{tabular}
\vspace{0.1cm}
\caption{\textbf{Comparison of supervision between our method and others.} The supervision in our work only comes from labeling the CAD models and their texture when generating the synthetic images. The second column indicates whether in-domain data (VOC \emph{train+val} set) is used in the methods. The amount of supervision used in each method increases from top to bottom.}
 \label{tab_compare}
 \vspace{-1cm}
\end{table*}

To summarize, this work contributes to the computer vision community in the following three aspects:
 \begin{itemize}
    \item we propose and implement a recognition framework that decomposes images into their shape and texture cues;
    \item we show that combining these cues improves classification and detection performance while using minimal supervision;
    \item we present a unified schema for learning from both web images and synthetic CAD images.
 \end{itemize}

\section{Related Work}
\label{related}

\noindent \textbf{Webly Supervised Learning.} The explosion of visual data on the Internet provides important sources of data for vision research. However, cleaning and annotating these data is costly and inefficient. Researchers have striven to design methods that learn visual representations and semantic concepts directly from the unlabeled data. Because the detection task requires stronger supervision than classification, most previous research work \cite{bergamo2010exploiting,fergus2010learning,li2010optimol,schroff2011harvesting} involving web images only tackles the object classification task. Some recent  work \cite{chen2013neil,divvala2014learning} aims at discovering common sense knowledge or capturing intra-concept variance. In the work of \cite{chen2015webly,divvala2014learning}, webly supervised object detectors are trained from image search results. We follow a similar approach as in~\cite{chen2015webly} to train our texture model from web search data, but also add shape information using a CAD-based CNN.

\vspace{0.1in}
\noindent \textbf{Utilization of CAD Models.} CAD models had been used by researchers since the early stages of computer vision. Recent work involving 3D CAD models focuses on pose prediction \cite{liebelt2010multi,stark2010back,su2015render,sun2009multi}. Other recent work applied CAD models to 2D object detection \cite{peng2015learning,sun2014virtual} by rendering synthetic 2D images from 3D CAD models and using them to augment the training data. The main drawback of these methods is that the rendered images are low-quality and lack real texture, which significantly hurts their performance. In contrast, we propose a two-stream architecture that adds texture information to the CAD-based shape channel. \cite{peng2015learning} explored several ways to simulate real images, by adding real-image background and textures onto the 3D models, but this requires additional human supervision to select appropriate background and texture images for each category. In this work, we propose more effective ways to simulate real data with less supervision.

\vspace{0.1in}
\noindent \textbf{Two-stream Learning.} The basic aim of two-stream learning is to model two-factor variations. \cite{tenenbaum2000separating} proposed a bilinear model to separate ``style" and ``content" of an image. In \cite{lin2015bilinear}, a two-stream architecture was proposed for fine-grained visual recognition, and the classifier is expressed as a product of two low-rank matrices. \cite{fragkiadaki2015learning,simonyan2014two} utilized two-stream architectures to model the temporal interactions and aspect of features. We propose a CNN-based two-stream architecture that learns intrinsic properties of objects from disparate data sources, with one stream learning to extract texture cues from real images and the other stream learning to extract shape information from CAD models. 

\vspace{-0.2cm}
\section{Approach}
\label{appro}
Our ultimate goal is to learn a good object classifier and object detector from the massive amount of visual data available via web search and from synthetic data. As illustrated in Figure \ref{fig:overview}, we introduce a two-stream learning architecture to extract the texture cues and shape information simultaneously. Each stream consists of three parts: the data acquisition component, the DCNN model and the object classifier or detector. The intuition is to utilize texture-oriented images from the web to train the texture stream and correspondingly, use shape-oriented images rendered from 3D CAD models to train the shape stream. 

\subsection{DCNN-based Two-Stream Model}
The history of two-stream learning can be traced back to over a decade ago when a ``bilinear model'' was proposed by \cite{tenenbaum2000separating} to separate the ``style'' and ``content'' of an image. More recent use~\cite{lin2015bilinear,fragkiadaki2015learning,simonyan2014two} of two-stream learning is based on a similar philosophy: employ different modalities to model different intrinsic visual properties, e.g. spatial features and temporal interactions. Inspired by this idea, we propose a two-stream learning architecture, with one stream modeling real image texture cues and the other modeling 3D shape information.  We demonstrate that the texture  and shape cues can reciprocally compensate for each other's errors through late fusion.

For fair comparison with other baselines, within each stream, we use the eight-layer ``AlexNet'' architecture proposed by \cite{alexnet}. It contains five convolutional layers, followed by two fully connected layers ($fc6$, $fc7$). After $fc7$, another fully connected layer ($fc8$) is applied to calculate the final class predictions. The network adopts ``dropout'' regularization to avoid overfitting and non-saturating neurons ($ReLU$ layers) to increases the nonlinear properties of the decision function and to speed up the training process. The network is trained by stochastic gradient descent and takes raw RGB image patches of size 227x227.

The last layer (\emph{fc8}) in each stream is represented by a softmax decision function. To combine the learned texture cues and shape information, we fuse the streams to render the final prediction as follows:
\begin{equation}
    P(I=j|\mathbf{x}) = \frac{e^{\mathbf{x}^{T}\mathbf{w}_{j}^{t}}}{2 \sum_{i=1}^{N} e^{\mathbf{x}^{T}\mathbf{w}_{i}^{t}}}+\frac{e^{\mathbf{x}^{T}\mathbf{w}_{j}^{s}}}{2 \sum_{i=1}^{N} e^{\mathbf{x}^{T}\mathbf{w}_{i}^{s}}} 
    \label{equ_fusion}
\end{equation}
where $P(I=j|\mathbf{x})$ denotes the probability that image $I$ belongs to category $j$ given feature vector $\mathbf{x}$ (\emph{fc7} feature in this case); $\mathbf{x}^{T}$, $N$, $\mathbf{w}_{i}^{t}$, $\mathbf{w}_{i}^{s}$ are the transpose of $\mathbf{x}$, the number of total categories, weight vector for category $i$ in Texture CNN, weight vector for category $i$ in Shape CNN, respectively.

The final probability $P(I=j|\mathbf{x})$ is used as the score for Two-Stream classifier and detector.

\subsection{Texture CNN Stream}
\label{sub:texture}

Previous work \cite{peng2015learning,viscnn} has shown that discriminative texture information is crucial for object classification and object detection systems. The challenge is how to obtain large scale accurate texture data with the least effort and how to prune the noisy images from unrelated search results. Previous approaches~\cite{chen2015webly,chen2013neil,divvala2014learning} have tried various search engines to form the texture bank, while other research work~\cite{fan2010harvesting,xia2014well} attempt to clean the data.

\noindent \textbf{Noise Data Pruning.} We assume the distribution of features of the higher CNN layers follows a multivariate normal distribution, thus we can fit the data from each class to the domain-specific Gaussian distribution as follows:

\begin{equation}
    f_{\mathbf{x}} (x_{1},x_{2} ... x_{k}) =   
    \frac{1}{\sqrt{(2\pi)^{k}|\sum|}}*exp(-\frac{1}{2}(\mathbf{x}-u)^T \sum{}{} ^{-1} (\mathbf{x}-u))
\end{equation}

where $\mathbf{x}$ is an k-dimensional feature vector and $\sum$ is the covariance matrix.

To remove outliers, for each category $j$ ($j\in [1,N]$, $N$ is the category number), we start from the downloaded image set $S_{j}$ with noise data and an empty set $T_{j}$. For each image $i$, we perform outlier removal by
 \begin{equation}
     T_{j}=\left\{\begin{matrix}
T_{j}\cup S_{j}(i), & P(S(i)=j|u^{j},\sum{}{}^{j}) \geqslant \varepsilon^{j} \\ 
 T_{j},& P(S(i)=j|u^{j},\sum{}{}^{j}) < \varepsilon^{j}
\end{matrix}\right.
\label{equ:gus}
 \end{equation}
where $P(S(i)=j|u,\sum)$, $\varepsilon^{j}, u^{j}, \sum{}{}^{j}$ are the probability that image $i$ belongs to category j, the pruning threshold for category $j$, the mean and covariance matrix of domain specific Gaussian distribution, respectively. We then use \{$T_{1}$, $T_{2}$, ..., $T_{N}$\} to train the Texture CNN.

\subsection{Shape CNN Model}
\label{appsub:shape}
Crowdsourced 3D models are easily accessible online and can be used to render unlimited images with different backgrounds, textures, and poses~\cite{su2015render,peng2015learning}. The widely used 3D Warehouse\footnote{https://3dwarehouse.sketchup.com}, Stanford Shapenet\footnote{http://shapenet.cs.stanford.edu/} provide numerous 3D CAD models for research use. Previous work has shown the great potential of synthetic images rendered from 3D CAD models for object detection~\cite{peng2015learning} and pose estimation~\cite{su2015render}. 

The flexibility and rigidity give 3D CAD models the unique merit for monitoring image properties such as background, texture and object pose, with constant shape information. However, the drawbacks of synthetic data are also obvious:
\begin{enumerate}
    \item \textbf{Lack of realism:} Sythetic images generally lack realistic intensity information which explicitly reflect fundamental visual cues such as texture, illumination, background.
    \item \textbf{Statistic Mismatch:} The statistics (eg. edge gradient) of synthetic image are different from realistic images. Thus the discriminatory information preserved by the DCNN trained on synthetic images may lose its effect on realistic images. 
\end{enumerate}

\begin{wrapfigure}{r}{0.5\textwidth}
\vspace{-1cm}
\begin{center}
    \includegraphics[width=0.48\textwidth]{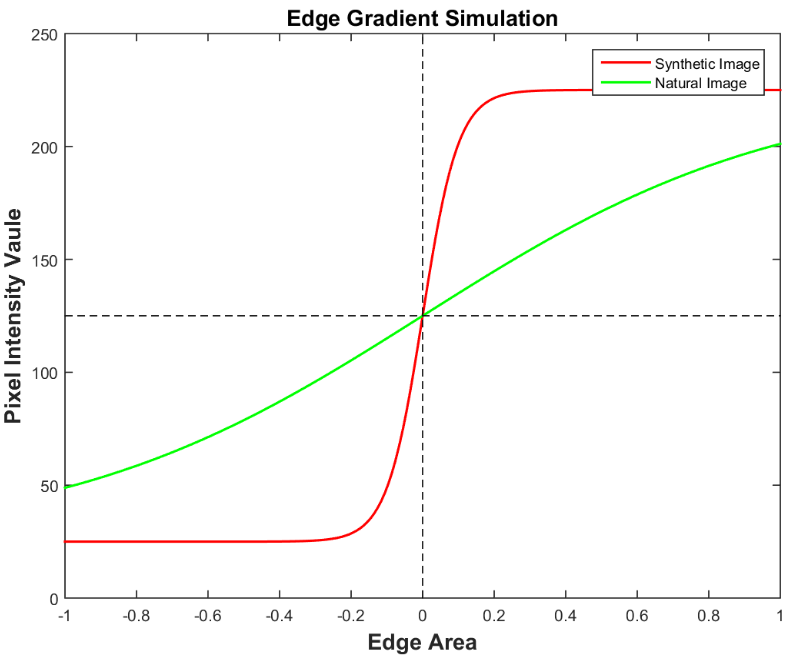}
  \end{center}
  \vspace{-0.7cm}
  \caption{\small{\textbf{Illustration of Edge Gradient}. Synthetic images rendered with white backgrounds tend have higher contrast edges around the outline of the object than natural images taken in the wild. Note that this figure is an illustration, not representing real pixel intensities.}}
   \vspace{-0.7cm}
   \label{fig:simulation}
\end{wrapfigure}
\noindent\textbf{Simulating Real-Image Statistics:} One flaw of synthetic images is that the instance is inconsistent with the background.
Thus, even rendered with very sophisticated parameters (pose variation, illumination, etc.), the statistics mismatch in intensity level still remains. Figure \ref{fig:simulation} illustrates the difference of edge gradients between synthetic images rendered with white backgrounds and real images. 

After analysing the synthetic data, we find the objects in the synthetic images tend to have higher contrast edges compared to real images taken in the wild. Adding more realistic backgrounds~\cite{peng2015learning} is a good way to decrease the contrast, but may obscure the object if the background is not chosen carefully. Instead, in this work, for each image $I$, we process $I$ by:
\begin{equation}
    I' = \psi_{\mathbf{G}}(I) + \xi_{G}
\end{equation}
where $\psi_{\mathbf{G}}(.)$ is a smoothing function based on a Gaussian filter and $\xi_{G}$ is a Gaussian noise generator. $\psi_{\mathbf{G}}(.)$ is used to mitigate the sharp edge contrast and $\xi_{G}$ to increase the intensity variations. 

\vspace{-0.2cm}

\section{Experiments}
\label{exper}
In this section, we  describe our experimental settings in detail. We start from downloading texture images via web search and rendering shape data from 3D CAD models. We evaluate our Two-Stream CNN classifier and detector on the standard benchmark PASCAL VOC 2007 \cite{pascal} dataset.
\vspace{-0.3cm}
\subsection{Data Acquisition}
\label{exp:data}
\noindent \textbf{Texture Data} As illustrated in Figure \ref{fig:overview}, we leverage a text-based image search engine (Google image search engine in our experiments) to collect the image data. Most of the images returned by Google contain a single object centered in the picture. This is good news for an algorithm attempting to learn the main features of a certain category. However, the drawback is the returned data is noisy and highly biased. For example, the top results returned by searching ``aeroplane'' may contain many toy aeroplane and paper plane images. To make matters worse, some returned images contain no ``aeroplane'', but objects from other categories.

We use the name of each object category as the query for Google image search engine to collect the training images. After removing unreadable images, there are about 900 images for each category and 18212 images in total. 

\begin{figure}[t]
   \centering
    
         \includegraphics[width=\textwidth]{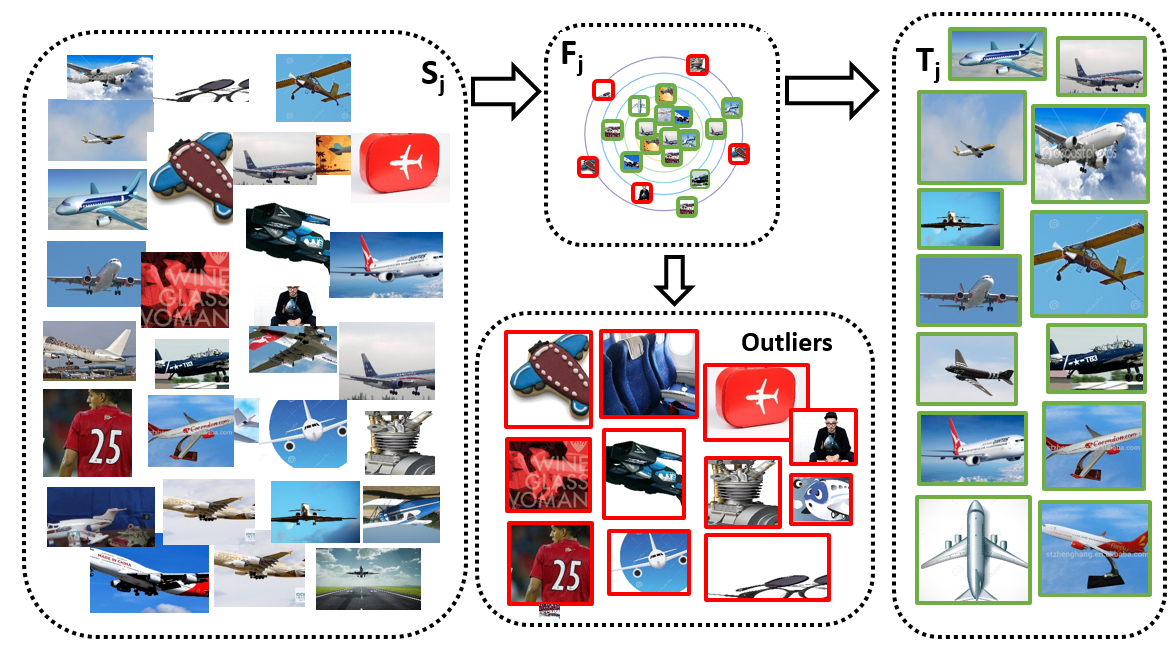}   
   \vspace{-0.2cm}
   \caption{\textbf{Illustration Of Noisy Data Pruning}. We fit the downloaded data to a multivariate normal distribution and remove the outliers if their probability is less than a learned threshold. (Best viewed in color) }
   \vspace{-0.5cm}
   \label{fig:purify}
\end{figure}

With millions of parameters, the CNN model easily overfits to small dataset. Therefore, data augmentation is valuable. Since most of the images are object-centered, we crop 40 patches by randomly locating the top-left corner $(x_{1}, y_{1})$ and bottom-right corner $(x_{2}, y_{2})$ by the following constraint: 
\begin{equation}
 \left\{\begin{matrix}
x_{1} \in[\frac{W}{20},\frac{3W}{20}],& y_{1} \in[\frac{H}{20},\frac{3H}{20}]\\ 
&\\
x_{2}\in[\frac{17W}{20},\frac{19W}{20}], & y_{2} \in[\frac{17H}{20},\frac{19H}{20}]

\end{matrix}\right.   
\end{equation}

$(W, H)$ are the width and height of original image. The constraint ensures that 49\%-81\% of the center area of the image is reserved. This image subsampling process leaves us about 0.7 million images to train the Texture CNN. 

We further utilize the approach illustrated in Section~\ref{sub:texture} to remove outliers from the downloaded data. For each category $j$ ($j\in[1,N]$, $N$ is the class number), we denote all the image patches after image subsampling as $S_{j}$. We adopt a DCNN architecture, known as ``AlexNet" to extract $fc7$ feature for each patch $i \in S_{j}$ to form the $fc7$ feature set $F_{j}$. We fit the $F_{j}$ to a multivariate normal distribution $\mathcal{N}(u, \sum)$ and compute the probability of each image path $i$. Through the fitting process we can find domain-specific variables $u^{j}$ and $\sum{}{}^j$. The threshold $\varepsilon^{j}$ in Equation \ref{equ:gus} is set so that the probabilities of 80\% of patches from $S_{j}$ are larger than it. Figure \ref{fig:purify} shows some samples which have been pruned out from the keyword search for ``aeroplane''.

\vspace{0.3cm}
\noindent \textbf{Shape Data} 3D CAD models of thousands of categories are available online. We utilize the 3D CAD models provided by \cite{peng2015learning} to generate our training images. These 3D CAD models were downloaded from 3D Warehouse by querying the name of the target categories. However, these models contain many ``outliers'' (eg. tire CAD models while searching car CAD models). To solve this, we manually selected the positive examples and delete the ``outliers''. To be consistent with our target dataset, we only adopt 547 3D CAD models for the 20 categories in PASCAL 2007, ranging from ``aeroplane'' to ``tv-monitor''.

\begin{figure}[t]
   \centering
         \includegraphics[width=\textwidth]{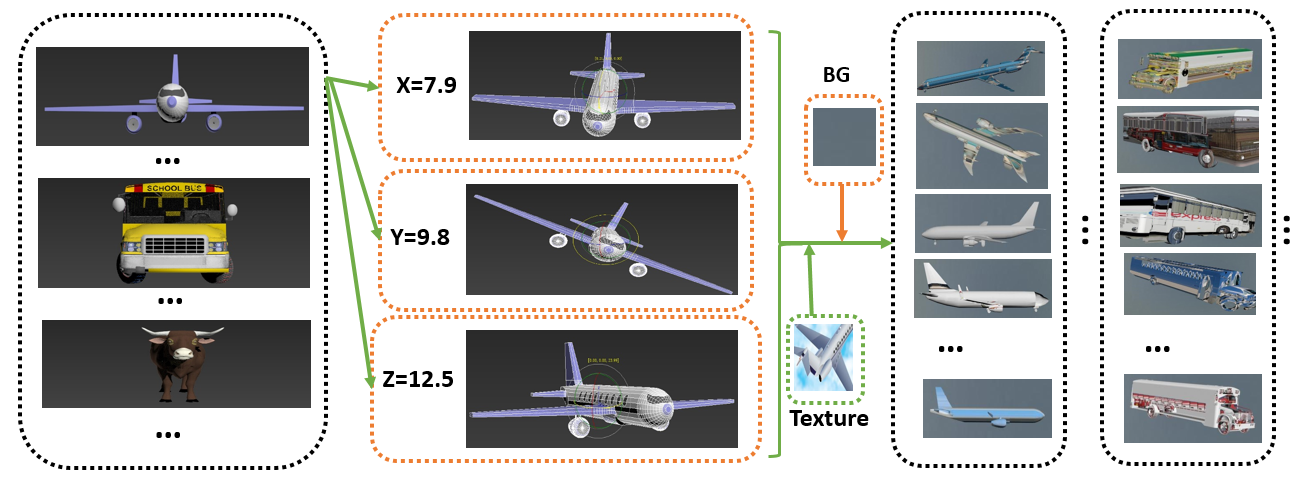}
   \vspace{-0.4cm}
   \caption{\textbf{Synthetic Data Rendering Process}. For each 3D CAD model, we first align the model to the front view and set rotation parameters $(X, Y, Z)$. The 3D CAD model is then rotated by the chosen parameters. To mitigate the intensity contrast around object edges, we add a background and texture to the final synthetic image.}
   \vspace{-0.5cm}
   \label{fig:syn}
\end{figure}

AutoDesk 3ds MAX\footnote{http://www.autodesk.com/store/products/3ds-max} is adopted to generate the synthetic images, with the entire generation process completed automatically by a 3ds Max Script. The rendering process is almost the same as \cite{peng2015learning} (we refer the reader to this work for more details), except that our approach  generates possible poses by exhaustively selecting $(X,Y,Z)$ rotation parameters, where $X, Y$ and $Z$ are the degree that the 3D CAD model needs to rotate around X-axis, Y-axis, Z-axis. As shown in figure \ref{fig:syn}, for each 3D CAD model, we first align the model to front view and set rotation parameters $(X, Y, Z)$. The 3D CAD model is then rotated by the chosen parameters. 
In our experiments, we only increment one variable from $(X,Y,Z)$ by 2 degrees at one time, constrained by $X,Y\in [-10, 10]$ and $Z\in [70,110]\cup [250,290]$ to cover possible intra-category variations. In total, we generate 833,140 synthetic training images to train our Shape CNN model.

To mitigate the intensity contrast around object edges, we add the mean image of ImageNet\cite{ImageNet} as the background to the final synthetic images. In our ablation study, we try either texture-mapping objects with real image textures as in~\cite{peng2015learning}, or using uniform gray (UG) texture. 

\begin{figure}[t]
   \centering
    
         \includegraphics[width=0.8\textwidth]{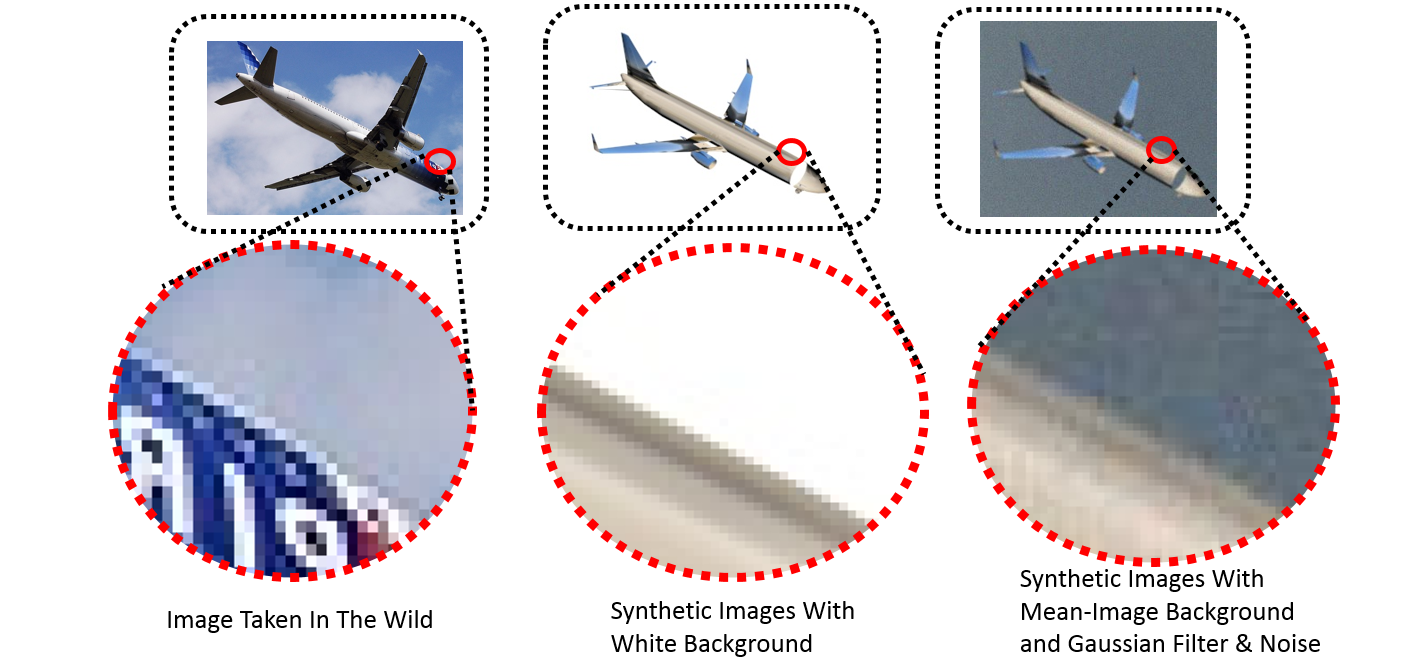}

   \caption{\textbf{Illustration of different edge gradients.} The edge gradients of synthetic images with white background differ drastically from images taken in the wild. In the third image, we replace the white background with a mean image computed from ImageNet\cite{ImageNet} images and apply  $\psi_{G}$(.) to smooth the edge contrast. We further add some Gaussian noise $\xi_{G}$ to the image to increase background variation.}
   \label{fig:edge}
\end{figure}

As addressed in Section~\ref{appsub:shape}, one shortcoming of synthetic data is the statistic mismatch, especially for the intensity contrast around the object edges. Figure \ref{fig:edge} shows the difference in edge gradient between synthetic images with white background and images taken in the wild.
Recent work on DCNN visualization \cite{zeiler2014visualizing} has shown that parameters in lower layers are more sensitive to edges. Thus matching the statistics of synthetic images to those of real images is a good way to decrease the domain shift. We apply a Gaussian filter based smoothing function $\psi_{G}$(.) for every synthetic image, and then add  Gaussian  noise $\xi_{G}$ to the smoothed images. In our experiment, we use $\mathcal{N}$(0, 1) for $\psi_{G}$(.) and $\mathcal{N}$(0, 0.01) for $\xi_{G}$.
The third image in Figure \ref{fig:edge} illustrates how the edge gradients become more similar to real image gradients after $\psi_{G}$(.) and $\xi_{G}$ are applied.

\subsection{Classification Results}

We evaluate our approach on standard benchmark PASCAL VOC 2007\cite{pascal}. PASCAL VOC dataset is originally collected for five challenges: classification, detection, segmentation, action classification and person layout. PASCAL VOC 2007 has 20 categories ranging from people, animals, plant to man-made rigid objects, and contains 5011 training/validation images and 4952 testing images. In our experiments, we only use testing images for evaluation.

Among the 4952 testing images, 14976 objects are annotated with tight bounding boxes. In our classification experiments, we crop 14976 patches (one patch for one object) with the help of these bounding boxes to generate test set. 

The DCNN in each stream is initialized with the parameters pre-trained on ImageNet\cite{ImageNet}. The last output layer is changed from a 1000-way classifier to a 20-way classifier and is randomly initialized with $\mathcal{N}(0, 0.01)$. The two DCNNs are trained with same settings, with the base learning rate to be 0.001, momentum to be 0.9 and weight decay to be 0.0005. Two dropout layers are adopted after $fc6,fc7$ with the dropout ratio to be 0.5.

The results in Table \ref{tab_cls} show when adding texture cues to Shape-CNN with Equation \ref{equ_fusion}, the performance rises from 28.3\%, 29.9\%, 31.7\% to 38.1\%, 38.7\%, 39.3\%, respectively. On the other side, adding shape cues to Texture-CNN can boost the performance from 36.7\% to 39.3\%, with $\psi_{G}$(.) and $\xi_{G}$ applied. The results also demonstrate that simulating the real statistics can benefit the classification results.

\begin{table*}[t]

\scriptsize

\noindent\begin{tabular}{ |p{2.0cm} | p{0.37cm}  p{0.37cm}  p{0.37cm}  p{0.37cm} p{0.37cm} p{0.37cm} p{0.37cm} p{0.37cm} p{0.37cm} p{0.37cm} p{0.37cm} p{0.37cm} p{0.37cm} p{0.37cm} p{0.37cm} p{0.37cm} p{0.37cm} p{0.37cm} p{0.37cm} p{0.37cm} |p{0.6cm}|    }

 \hline                       
  Method  & aer & bik & bir & boa & bot & bus & car & cat & chr & cow & tbl & dog & hrs & mbk & prs & plt & shp & sof & trn & tv & All\\
\hline
T-CNN& \textbf{61}& \textbf{76}& \textbf{36}& 39& 12& 44& \textbf{53}& 68& 12& 52& \textbf{16}& \textbf{46}& \textbf{58}& 56& 19& \textbf{86}& \textbf{63}& 22& \textbf{76}& \textbf{89}&36.7\\

\hline
\hline

S-CNN& 37& 24& 18& 50& \textbf{60}& \textbf{82}& 36& 54& 16& 78& 2& 10& 3& 56& 19& 31& 6& 26& 7& 77&28.3\\
S-CNN-G& 31& 20& 21& 58& 44& 80& 44& 42& 20& 73& 3& 9& 3& 66& 21& 42& 8& 34& 6& 80&29.9\\
S-CNN-GN& 32& 20& 20& 51& 53& 80& 47& 47& \textbf{25}& 68& 2& 12& 5& \textbf{69}& 21& 53& 8& \textbf{37}& 3& 85&31.7\\
\hline

F-CNN& 52& 66& 30& 50& 42& 74& 45& \textbf{70}& 16& \textbf{84}& 13& 33& 41& 59& 24& 71& 53& 27& 50& \textbf{89}&\textbf{38.1}\\
F-CNN-G& 56& 65& 31& \textbf{59}& 27& 76& 48& 63& 18& 77& 14& 35& 46& 64& \textbf{25}& 71& 55& 30& 51& \textbf{89}&\textbf{38.7}\\
F-CNN-GN& 56& 65& 31& 48& 32& 74& 51& 65& 21& 68& 13& 35& 47& 67& \textbf{25}& 77& 54& 31& 46& 90&\textbf{39.3}\\

\hline
\hline

S-CNN-UG&30&25&16&51&20&80&40&11&13&49&1&5&16&76&27&64&1&24&35&46&27.2\\
S-CNN-G-UG&25&16&15&43&20&86&43&8&15&57&1&3&5&81&21&53&1&30&14&38&27.5\\
S-CNN-GN-UG&35&23&13&55&33&82&37&9&22&34&0&4&3&78&29&56&0&29&19&39&29.4\\

\hline

F-CNN-UG&56&68&30&52&16&70&50&61&14&57&12&38&49&70&28&83&55&23&68&81&39.3\\
F-CNN-G-UG&56&59&29&48&16&79&50&62&15&61&11&35&46&78&26&80&54&26&56&81&38.1\\
F-CNN-GN-UG&62&65&28&55&20&74&49&61&19&50&13&38&48&75&32&81&56&26&62&84&41.4\\

\hline
\end{tabular}
\vspace{0.2cm}
\caption{\textbf{Classification Results}. Prefix ``S-'', ``T-'' and ``F-'' denote ``Shape'', ``Texture'', ``Fusion'', respectively. Suffix ``-G'', ``-N'', ``-UG'' indicate synthetic data is smoothed with $\psi_{G}$(.), is colored with Gaussian noise $\xi_{G}$, and is generated with uniform gray texture. The results show Fusion-CNN model outperforms Shape-CNN model and Texture-CNN model and simulating real statistics benefit object classification system with minimal supervision. Note the last column is not mean accuracy, but accuracy over all test set.}
 \label{tab_cls}
\vspace{-0.7cm}
\end{table*}

\begin{figure}[t]
    \centering

            \includegraphics[width=0.4\textwidth]{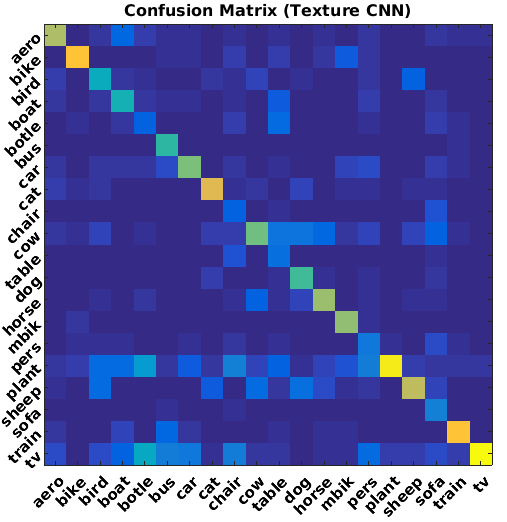}
            \includegraphics[width=0.4\textwidth]{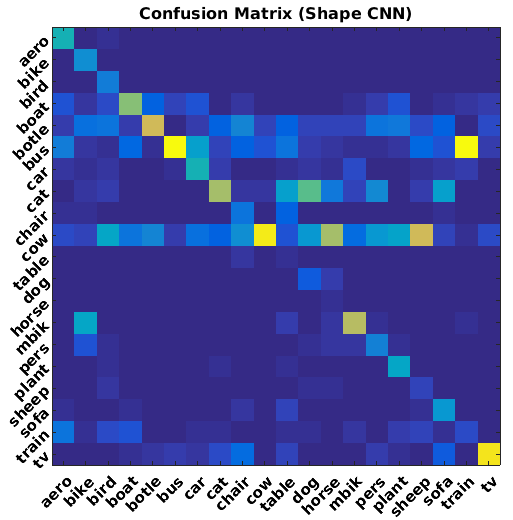}
            \includegraphics[width=0.4\textwidth]{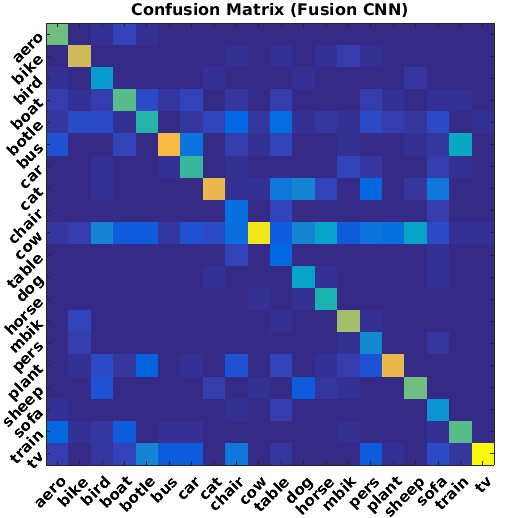}
            \includegraphics[width=0.4\textwidth]{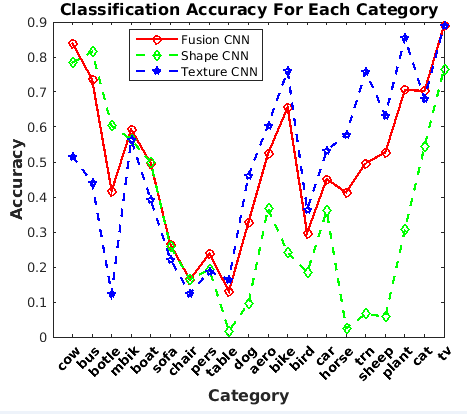}

    \caption{\textbf{Confusion matrix and classification results}. The confusion matrix has been normalized by the number of total images per category. From up to bottom, left to right, the four figures are: confusion matrix of texture CNN, confusion matrix of shape CNN, confusion matrix of fusion CNN, classification accuracy for each category. (Best viewed in color!) }
    \label{fig:confusion}
    \vspace{-0.4cm}
\end{figure}

To better analyze how texture cues and shape information compensate for each other, we plot the confusion matrix (with X-axis representing ground truth labels and Y-axis representing predictions) for Texture-CNN, Shape-CNN, Fusion-CNN in Figure \ref{fig:confusion} (Networks with ``-G'' or ``-GN" have similar results). For the convenience of comparison, we re-rank the order of categories and plot per-category accuracy in Figure \ref{fig:confusion}. There are some interesting findings:
\begin{itemize}
    \item The top-right confusion matrix (for Shape-CNN) in Figure \ref{fig:confusion} shows that CNN trained on synthetic images mistakes most of the ``train'' images for ``bus'', and mistakes ``horse'' and ``sheep'' images for ``cow'', which is not very surprising because they share similar shape visual information.
    \item From two confusion matrices on the top, we can see that Texture-CNN trained on web images tends to mistake other images for ``plant'' and ``TV'', while Shape-CNN is keen on categories like ``cow'', ``bus'', ``bottle'' etc.
    \item The last sub-figure in Figure \ref{fig:confusion} is per-category accuracy. We re-rank the order of categories for inspection convenience. Shape-CNN (green line) tends to perform well for the categories presented on the left and Texture-CNN (blue line) is more likely to get a high performance for the categories presented on the right. Taking a closer look at the categories, we find that Shape-CNN will work well for shape-oriented or rigid categories, e.g. bus, bottle, motorbike, boat, etc. Inversely, Texture-CNN is more likely to obtain higher performance on texture-oriented categories, such as cat, sheep, plant, horse, bird etc. The performance of Fusion-CNN is mostly between Shape-CNN and Texture-CNN and  never gets a very poor result, which is why it can work better than CNNs based on single cues. We also tried performing a max fusion over the two streams, but the performance improvement is not comparable with average fusion.
\end{itemize}

We also try removing the texture on the object from the synthetic image by replacing it with uniform gray (UG) pixels. As shown in Table \ref{tab_cls}, this achieves similar results, indicating that synthetically adding real texture to CAD models may not be important for this classification task.

\subsection{Object Detection Results}

In our detection experiments, we find ``S-CNN'' (``F-CNN''), ``S-CNN-G'' (``F-CNN-G'') and ``S-CNN-GN'' (``F-CNN-GN'') get comparable results, thus we only report result for ``S-CNN'' (``F-CNN'') in this section.

For detection, we followed the standard evaluation schema provided by \cite{pascal}: a prediction bounding box P is considered to be a valid detection if and only if the area of overlap $IoU$ exceeds 0.5. The $IoU$ is denoted with the following formula: $IoU$ = $\frac{area(B_{p}\cap B_{gt})}{area(B_{p} \cup B_{gt})}$, where $B_{p} \cap B_{gt}$ denotes the intersection of the predicted bounding box and the ground truth bounding box and $B_{p} \cup B_{gt}$ their union.

\vspace{0.2cm}

\noindent \textbf{Region Proposal} An excellent region proposal method contributes to the performance of both supervised and unsupervised learning. In our experiments, we adopt EdgeBox\cite{zitnick2014edge} to generate region proposals. EdgeBox is a efficient region proposal algorithm which generates bounding box proposals from edge maps obtained by contour detector. The bounding boxes are scored by the number of enclosed contours inside the boxes.   

\vspace{0.2cm}
Like in the classification task, for each region proposal, we pass it to Shape-CNN and Texture-CNN simultaneously, and fuse the last layers' activations. Similar to \cite{chen2015webly}, we randomly crop patches from YFCC\cite{thomee2015new} as the negative samples. Further, we  follow the schema of R-CNN\cite{RCNN} to compute mAP. We compare our method to following baselines.

\begin{itemize}
 \item \textbf{VCNN(ICCV'15)\cite{peng2015learning}} In this work, the authors propose to render domain-specific synthetic images from 3D CAD models and train an R-CNN\cite{RCNN} based object detector. Some results may involve minor supervision, e.g. selecting background and texture. We compare to their W-RR model (white background, real texture) where the amount of supervision is almost the same as in this work.
    \item \textbf{LEVAN(CVPR'14)\cite{divvala2014learning}} LEVAN uses items in Google N-grams as queris to collect training images from Internet. They propose a fully-automated approach to organize the visual knowledge about a concept and further apply their model to detection task on PASCAL VOC 2007.
    \item \textbf{Webly Supervised Object Detection(ICCV'15)\cite{chen2015webly}} The webly supervised learning approach collects images from Google and Flickr by searching for the name of a certain category and utilizes Examplar-LDA~\cite{hariharan2012discriminative} and agglomerative clustering~\cite{chen2014enriching} to generate the potential ``ground truth" bounding box. For fair comparison, we only compared to their results of images downloaded from Google.
   
     \item \textbf{Concept Learner(CVPR'15)\cite{zhou2015conceptlearner}} Concept learner is designed to discover thousands of visual concepts automatically from webly labeled images. It first trains a concept learner on the SBU dataset and selects the learned concept detectors to compute the average precision.
\end{itemize}

Results listed in Table \ref{tab_det} demonstrate that combining real-image texture information with Shape-CNN will boost the mAP from 15.0 to 19.7, a relative 31.3\% increase! Inversely, adding shape information to Texture-CNN boost the mAP from 18.1 to 19.7, which shows texture cues and shape information can compensate for each other in detection task. Despite the minimal amount of required supervision, our  Fusion-CNN also obtains higher performance than a purely 3D CAD model based method like \cite{peng2015learning}, a webly supervised approach like \cite{divvala2014learning} and a weakly supervised method where in-domain training data from PASCAL VOC 2007 is available \cite{divvala2014learning}. The results show that Fusion-CNN outperforms DCNN based on single visual cues and other methods where similar or higher levels of supervision are adopted. 

As an ablation study, we perform the same experiments on synthetic images generated without texture (``S-CNN-UG'', ``F-CNN-UG'' in table \ref{tab_det}). The results reveal that, unlike classification, adding some texture into the synthetic images helps to boost performance for the detection task.

\begin{table*}[t]
\scriptsize
\noindent\begin{tabular}{ p{2.4cm} |p{0.35cm}  p{0.35cm}  p{0.35cm}  p{0.35cm} p{0.35cm} p{0.35cm} p{0.35cm} p{0.35cm} p{0.35cm} p{0.35cm} p{0.35cm} p{0.35cm} p{0.35cm} p{0.35cm} p{0.35cm} p{0.35cm} p{0.35cm} p{0.35cm} p{0.35cm} p{0.35cm}  |p{0.6cm}   }
 \hline                       
  Method  & aer & bik & bir & boa & bot & bus & car & cat & chr & cow & tbl & dog & hrs & mbk & prs & plt & shp & sof & trn & tv & mAP\\
\hline

T-CNN & 22& 20& 19& 18& 9& 35& 28& \textbf{21}& 9& 13&\textbf{4}& 16&\textbf{29}& 31& 6& \textbf{11}& 15& 11& \textbf{25}& 19& 18.1\\
S-CNN & 20& 18& 18& 15& 9& 29& 23& 4& 9& 16& 0& 13& 19& 26& 13& 9& 14& 12& 4& 27& 15.0\\
F-CNN & 29& 23& \textbf{19}& \textbf{22}& 9& \textbf{41}& 29& 17& 9& \textbf{21}& 1& \textbf{20}& 23& \textbf{33}& 9& 9& \textbf{17}& \textbf{13}& 16& 27&\textbf{19.7}\\

\hline

S-CNN-UG & 22&17&13&12&9&23&26&2&2&13&0&6&11&29&5&9&2&10&1&17&11.4\\
F-CNN-UG &25&18&15&18&10&38&30&14&3&18&1&17&20&31&8&10&12&12&11&19&17\\

\hline
\hline

VCNN \cite{peng2015learning} & \textbf{36}& \textbf{23}& 17& 15& \textbf{12}& 25& \textbf{35}& 21& \textbf{11}& 16& 0.1& 16& 16& 29& \textbf{13}& 9& 4& 10& 0.6& \textbf{29}& 17\\
\hline
\hline
Levan, Webly\cite{divvala2014learning}& 14& 36& 13& 10& 9& 35& 36& 8& 10& 18& 7& 13& 31& 28& 6& 2& 19& 10& 24& 16& 17.2\\
Chen's Webly\cite{chen2015webly}& 35& 39& 18& 15& 8& 31& 39& 20& 16& 13& 15& 4& 21& 34& 9& 17& 15& 23& 28& 19& 20.9\\
\hline
\hline
Zhou's, Selected\cite{zhou2015conceptlearner}& 30& 34& 17& 13& 6& 44& 27& 23& 7& 16& 10& 21& 25& 36& 8& 9& 22& 17& 31& 18& 20.5\\
\hline
\hline
Siva's, Weakly\cite{siva2011weakly}& 13& 44& 3& 3& 0& 31& 44& 7& 0& 9& 10& 2& 29& 38& 5& 0& 0& 4& 34& 0& 13.9\\
Song's, Weakly\cite{song2014learning}& 8& 42& 20& 9& 10& 36& 39& 34& 1& 21& 10& 28& 29& 39& 9& 19& 21& 17& 36& 7& 22.7\\
\hline
\hline
RCNN, Full\cite{RCNN}& 58& 58& 39& 32& 24& 51& 59& 51& 20& 51& 41& 46& 52& 56& 43& 23& 48& 35& 51& 57& 44.7\\
\hline
\end{tabular}
\vspace{0.2cm}
\caption{\textbf{Illustration Of Detection Results}. Methods are categorized by their supervision type. ``Webly'', ``Selected'', ``Weakly'', ``Full'' represent webly supervision, selected supervision, weak supervision, full supervision, respectively. The definitions of these supervisions are the same as in \cite{zhou2015conceptlearner}. The supervision in our work only comes from labeling the CAD models and choosing proper texture for the CAD models when generating synthetic images. The supervision used in \cite{peng2015learning} is almost the same as in our approach, except that they also labeled pose for the CAD models. The results demonstrate that our Fusion-CNN model outperforms methods based on single visual cues and other methods with similar or higher required supervision effort.}
 \label{tab_det}
 \vspace{-0.6cm}

\end{table*}

\vspace{-0.4cm}

\section{Conclusion}
\label{con}

In this work, we proposed and implemented a novel minimally-supervised learning framework that decomposes images into their shape and texture and further demonstrated that texture cues and shape information can reciprocally compensate for each other. Furthermore, a unified learning schema, including pruning noise web data and simulating statistics of real images is introduced, both for web image based learning and synthetic image based learning. Finally, our classification and detection experiments on VOC 2007 show that our Fusion-CNN with minimal supervision outperforms DCNNs based on single cues (only shape, only texture) and previous methods that require similar or more supervision effort. We believe our model is valuable for scaling recognition to many visual object categories and can be generalized to other generic tasks such as pose detection, robotic grasping and object manipulation.

\vspace{0.2cm}


\noindent {\bf Acknowledgement}. This research was supported by NSF award IIS-1451244 and a generous donation from the NVIDIA corporation.


\section{Appendix}
\begin{figure}
    \centering

    \includegraphics[width=0.24\textwidth]{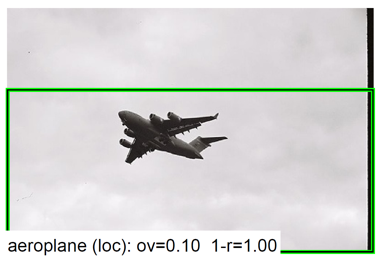}
    \includegraphics[width=0.24\textwidth]{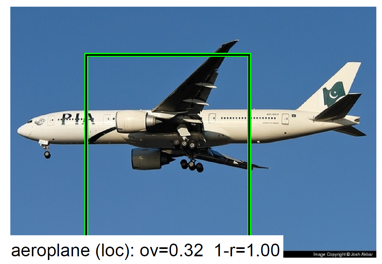}
    \includegraphics[width=0.24\textwidth]{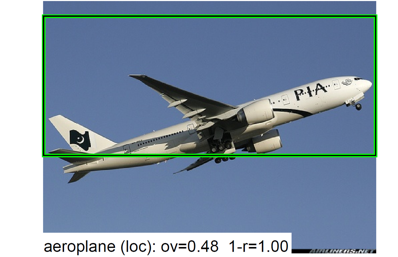}
    \includegraphics[width=0.24\textwidth]{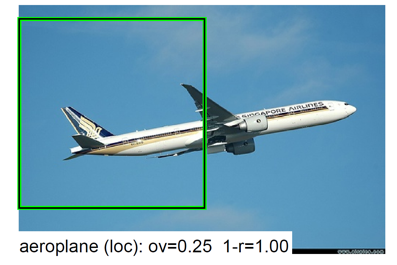}
    \includegraphics[width=0.24\textwidth]{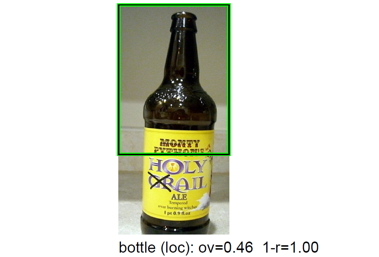}
    \includegraphics[width=0.24\textwidth]{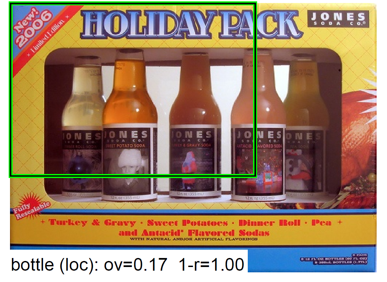}
    \includegraphics[width=0.24\textwidth]{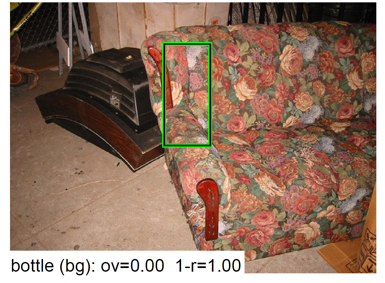}
    \includegraphics[width=0.24\textwidth]{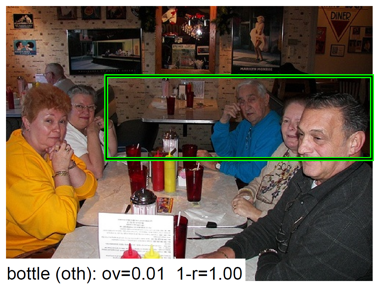}
    \includegraphics[width=0.24\textwidth]{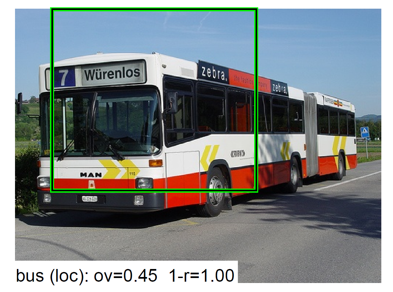}
    \includegraphics[width=0.24\textwidth]{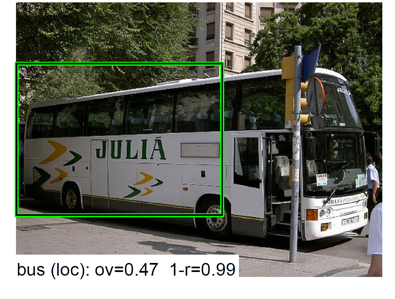}
    \includegraphics[width=0.24\textwidth]{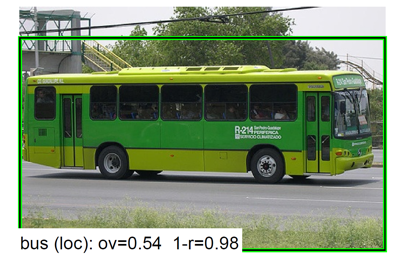}
    \includegraphics[width=0.24\textwidth]{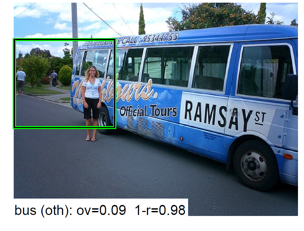}
       \includegraphics[width=0.24\textwidth]{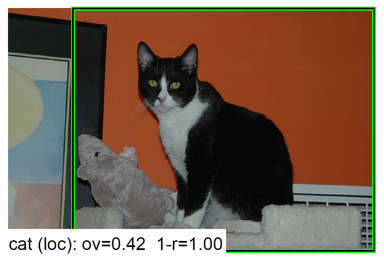}
    \includegraphics[width=0.24\textwidth]{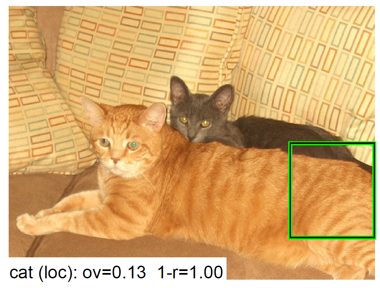}
    \includegraphics[width=0.24\textwidth]{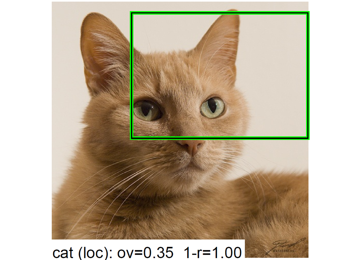}
    \includegraphics[width=0.24\textwidth]{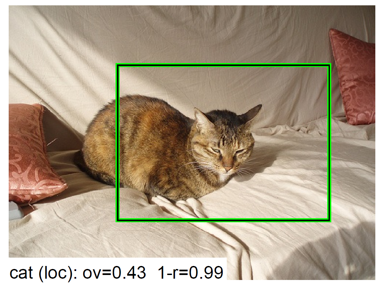}
       \includegraphics[width=0.24\textwidth]{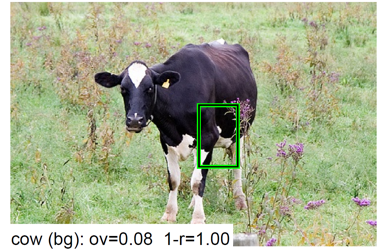}
    \includegraphics[width=0.24\textwidth]{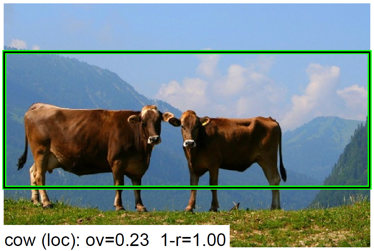}
    \includegraphics[width=0.24\textwidth]{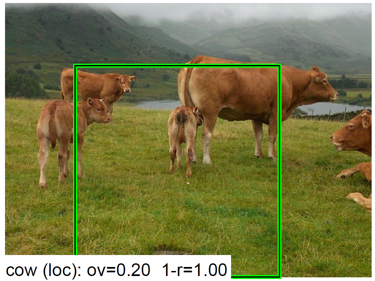}
    \includegraphics[width=0.24\textwidth]{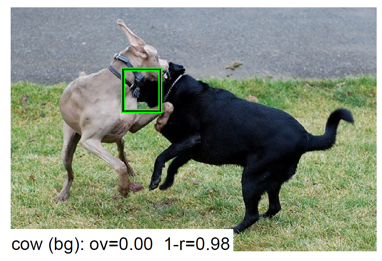}
       \includegraphics[width=0.24\textwidth]{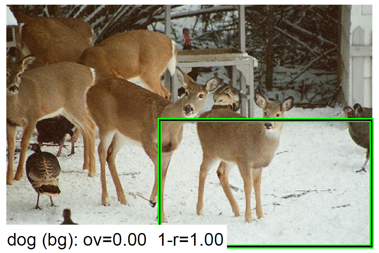}
    \includegraphics[width=0.24\textwidth]{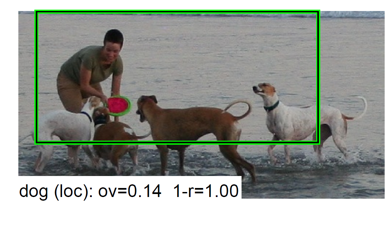}
    \includegraphics[width=0.24\textwidth]{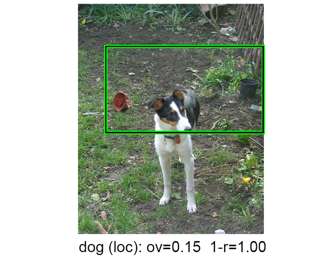}
    \includegraphics[width=0.24\textwidth]{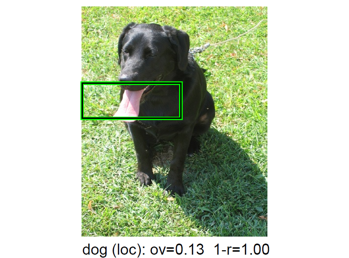}
       \includegraphics[width=0.24\textwidth]{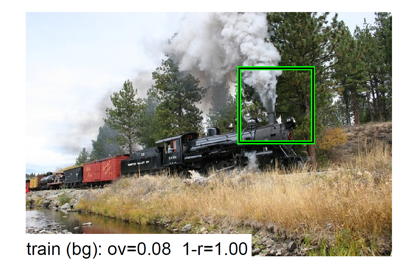}
    \includegraphics[width=0.24\textwidth]{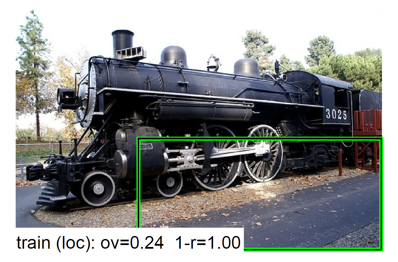}
    \includegraphics[width=0.24\textwidth]{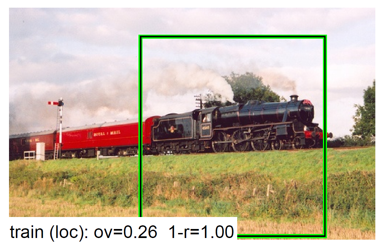}
    \includegraphics[width=0.24\textwidth]{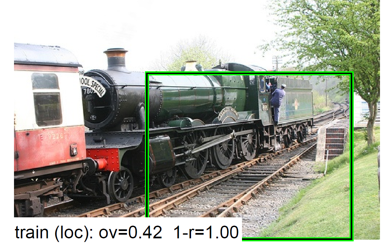}
    \caption{\textbf{Examples of top false positives.} We show the top four false positives for \textbf{F-CNN} ``aeroplane'', ``bottle'', ``bus'', ``cat'', ``cow'', ``dog'' and ``train'' detectors. The text on the bottom of each image shows the type of error, the amount of overlap (``ov'') with a ground truth object, and the fraction of true positives that are ranked after the given false positive (``1-r'', for 1-recall) }

    \label{fig_topfp}
\end{figure}

\noindent {\bf A. Detection Result Diagnosis}\\
We further use the diagnosis tools provided in~\cite{hoiem-analysis} to analyze our detectors, Table~\ref{tab_diag} and Figue~\ref{fig_topfp} highlight some of the interesting observations.

Table \ref{tab_diag} illustrates the diagnosis results. From top to bottom, each row represents diagnosis for ``\textbf{animals}'' (including ``birds'', ``cat'', ``cow'', ``dog'', ``horse'', ``person'', ``sheep''), ``\textbf{vehicles}'' (including ``aeroplane'', ``bike'',``boat'', ``bus'', ``car'', ``motorbike'', ``train'') and ``\textbf{furniture}'' (including ``chair'', ``table'', ``sofa'').
The diagnosis tools \cite{hoiem-analysis} will categorize the false positive samples into four categories: \textbf{Localization error} (\textbf{Loc}), \textbf{Confusion with similar objects} (\textbf{Sim}), \textbf{Confusion with dissimilar objects} (\textbf{Oth}), \textbf{Confusion with background} (\textbf{BG}). (We refer the reader to \cite{hoiem-analysis} for more details). After analyzing the distribution of these four type of errors, we list some interesting observations:
\begin{itemize}
    \item Localization error (\textbf{Loc}, blue area in the figures) is the majority of the false positives for \textbf{T-CNN}, mainly because training images (downloaded from Google image search engine) for Texture-CNN do not provide ground truth bounding boxes.  Other issues like multiple instances in one bounding box also cause localization error. 
    \item For \textbf{S-CNN}, confusion with similar objects (\textbf{Sim}, red area in the figures) accounts for a large portion of the false positives, especially for ``animals''. The synthetic data used to train Shape-CNN lacks discriminative texture cues. For example, the rendered sheep images, horse images and cow images share the same visual representations in shape.
    \item Compared to ``anmimals'', ``vehiches'' will have less confusion with similar objects (\textbf{Sim}) errors and confusion with dissimilar objects (\textbf{Oth}) errors; on the contrary, ``furniture'' will have more.  
    \item The distribution of false positives for \textbf{F-CNN} is a compromise of \textbf{T-CNN} and \textbf{S-CNN}.
\end{itemize}

\begin{table}

\begin{center}
\begin{tabular}{ |c|c|c| } 
\hline
\textbf{T-CNN}&\textbf{S-CNN}&\textbf{F-CNN}\\

 \hline
  \includegraphics[width=0.31\textwidth]{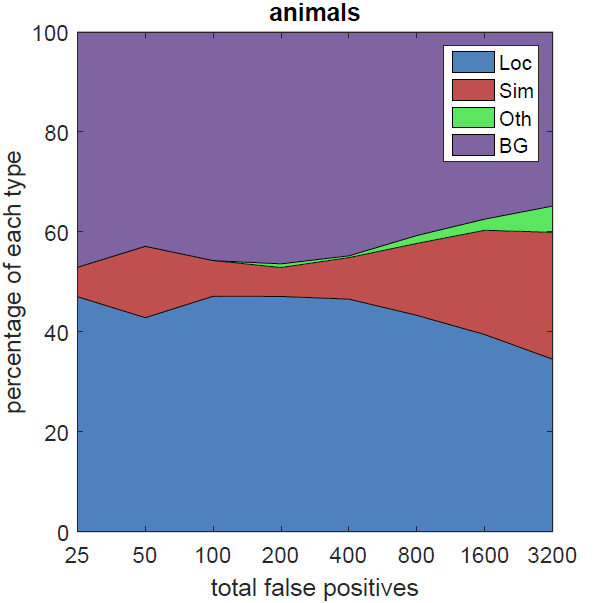} &
 \includegraphics[width=0.31\textwidth]{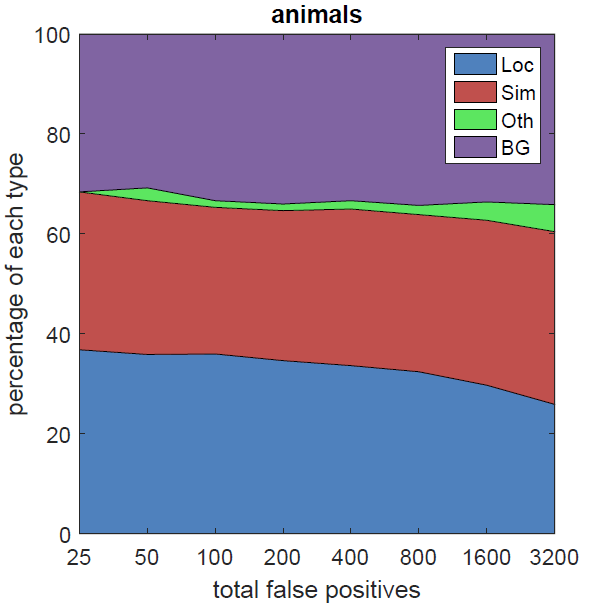} & 
 \includegraphics[width=0.31\textwidth]{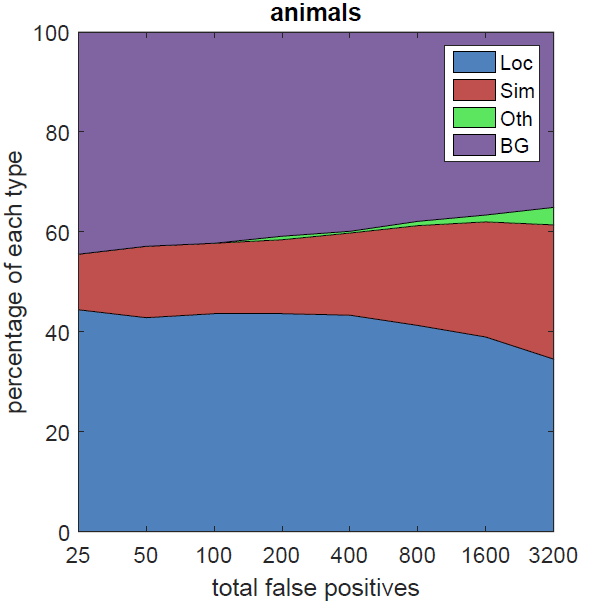} \\ 
 \hline
 \includegraphics[width=0.31\textwidth]{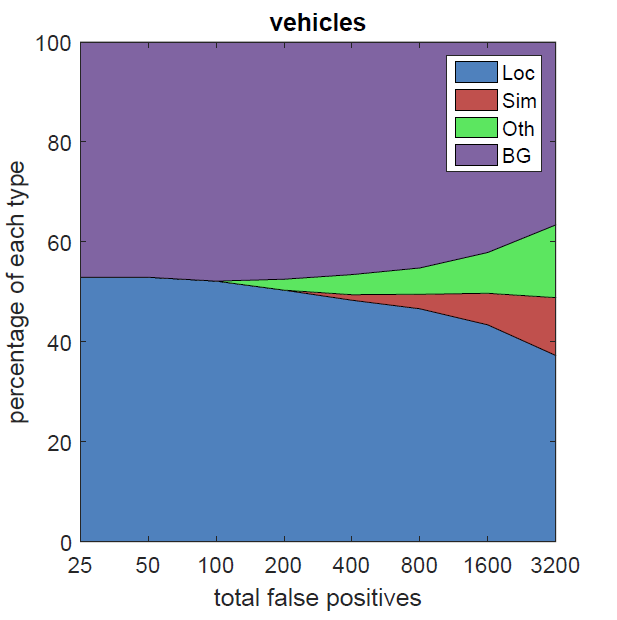} &
 \includegraphics[width=0.31\textwidth]{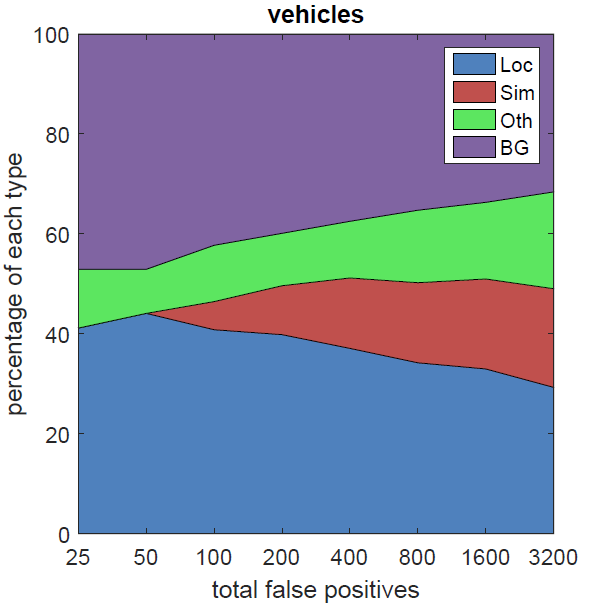} & 
 \includegraphics[width=0.31\textwidth]{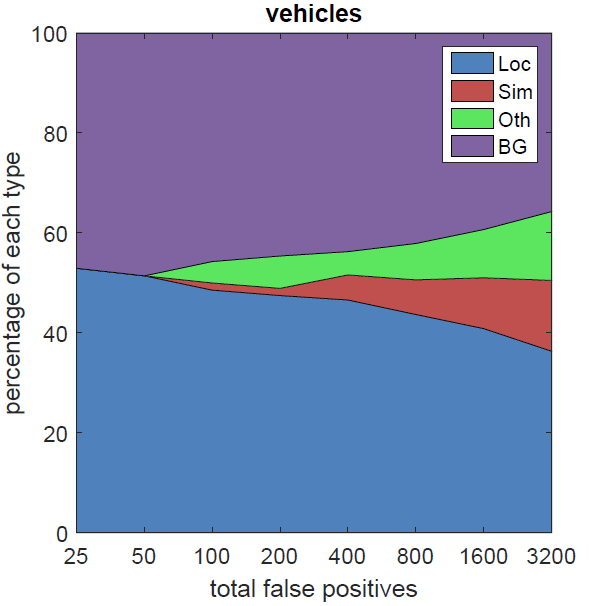} \\ 
 \hline
 \includegraphics[width=0.31\textwidth]{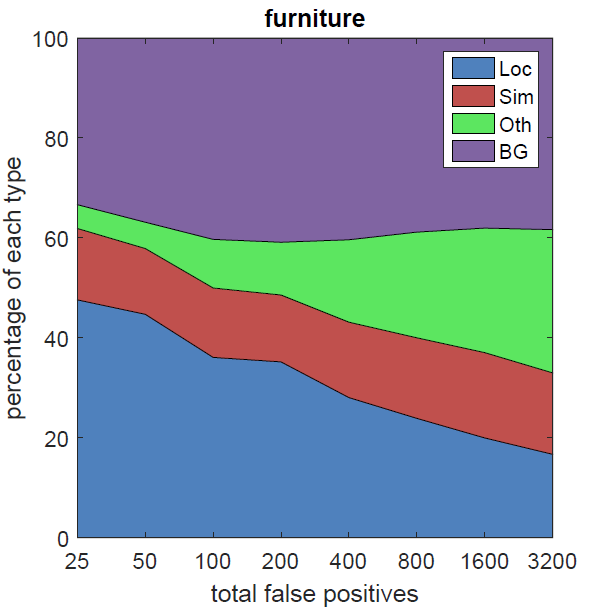} &
 \includegraphics[width=0.31\textwidth]{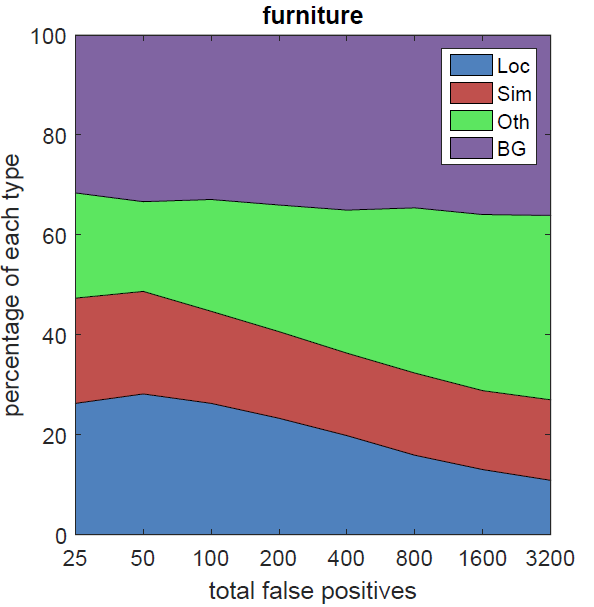} & 
 \includegraphics[width=0.31\textwidth]{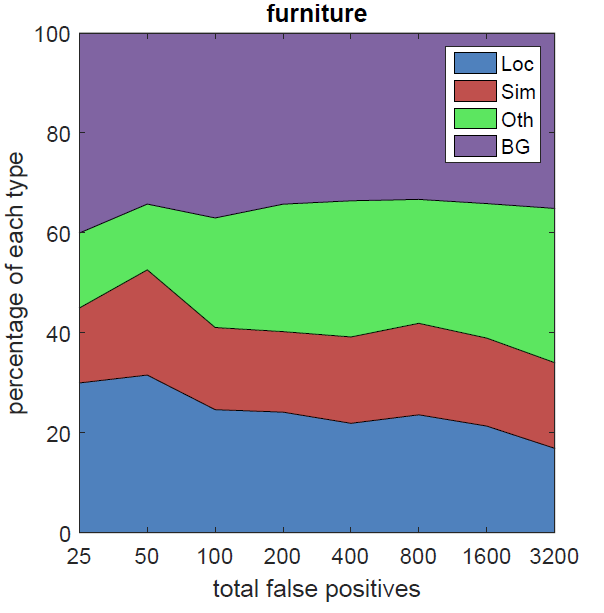} \\ 
 \hline
\end{tabular}
\end{center}

\caption{\textbf{Illustration of diagnosis.} We visualize the failure model of our models. From up to bottom, each row shows the diagnosis for ``animals'', ``vehicles'' and ``furniture''. The colored area indicates the percentage of false positives of each failure type. (Best viewed in color!)}
\label{tab_diag}
\end{table}

\noindent {\bf  B. Examples Of Top False Positives}\\

In figure \ref{fig_topfp}, we show the top four false positives for category ``aeroplane'', ``bottle'', ``bus'', ``cat'', ``cow'', ``dog'' and ``train''. These figures demonstrate that localizing the objects precisely is still a great challenge for object detector with little supervision. 
\vspace{2cm}

\bibliographystyle{splncs}

\begin{thebibliography}{1}


\bibitem{bergamo2010exploiting}
Bergamo, Alessandro and Torresani, Lorenzo:
Exploiting weakly-labeled web images to improve object classification: a domain adaptation approach.
Advances in Neural Information Processing Systems
(2010) 181--189

\bibitem{boser1992training}
Boser, Bernhard E and Guyon, Isabelle M and Vapnik, Vladimir N:
A training algorithm for optimal margin classifiers.
Proceedings of the fifth annual workshop on Computational learning theory
(1992) 144-152

\bibitem{chen2015webly}
Chen, Xinlei and Gupta, Abhinav:
Webly supervised learning of convolutional networks.
Proceedings of the IEEE International Conference on Computer Vision 
(2015) 1431--1439

\bibitem{chen2013neil}
Chen, Xinlei and Shrivastava, Abhinav and Gupta, Abhinav:
Neil: Extracting visual knowledge from web data.
Proceedings of the IEEE International Conference on Computer Vision
(2013) 1409--1416

\bibitem{chen2014enriching}
Chen, Xinlei and Shrivastava, Abhinav and Gupta, Abhinav:
Enriching visual knowledge bases via object discovery and segmentation.
Proceedings of the IEEE Conference on Computer Vision and Pattern Recognition
(2014) 2027--2034

\bibitem{dalal2005histograms}
Dalal, N. and Triggs, B.:
Histograms of oriented gradients for human detection.
Proc. IEEE Conf. Computer Vision and Pattern Recognition (2005)

\bibitem{ImageNet}
Deng, Jia and Dong, Wei and Socher, Richard and Li, Li-Jia and Li, Kai and Fei-Fei, Li:
Imagenet: A large-scale hierarchical image database.
IEEE Conference on Computer Vision and Pattern Recognition
(2009) 248--255

\bibitem{divvala2014learning}
Divvala, Santosh and Farhadi, Ali and Guestrin, Carlos:
Learning everything about anything: Webly-supervised visual concept learning.
Proceedings of the IEEE Conference on Computer Vision and Pattern Recognition
(2014) 3270--3277

\bibitem{hoiem-analysis}
D. Hoeim and Y. Chodpathumwan and Q. Dai:
Diagnosing Error in Object Detectors.
In Proc. ECCV (2012)

\bibitem{pascal}
Everingham, M. and Van~Gool, L. and Williams, C. K. I. and Winn, J. and Zisserman, A.:
The Pascal Visual Object Classes (VOC) Challenge.
International Journal of Computer Vision
\textbf{88} (2010) 303--338


\bibitem{fan2010harvesting}
Fan, Jianping and Shen, Yi and Zhou, Ning and Gao, Yuli:
Harvesting large-scale weakly-tagged image databases from the web.
CVPR (2010) 802--809


\bibitem{lsvm-pami}
Felzenszwalb, P. F. and Girshick, R. B. and McAllester, D. and Ramanan, D.:
Object Detection with Discriminatively Trained Part Based Models.
IEEE Transactions on Pattern Analysis and Machine Intelligence
\textbf{32} (2010) 1627--1645


\bibitem{fergus2010learning}
Fergus, Rob and Fei-Fei, Li and Perona, Pietro and Zisserman, Andrew:
Learning object categories from internet image searches.
Proceedings of the IEEE
\textbf{98} (2010) 1453--1466


\bibitem{fragkiadaki2015learning}
Fragkiadaki, Katerina and Arbelaez, Pablo and Felsen, Panna and Malik, Jitendra:
Learning to segment moving objects in videos.
Computer Vision and Pattern Recognition (CVPR), 2015 IEEE Conference on
(2015) 4083--4090


\bibitem{RCNN}
Girshick, Ross and Donahue, Jeff and Darrell, Trevor and Malik, Jitendra:
Rich feature hierarchies for accurate object detection and semantic segmentation
arXiv preprint arXiv:1311.2524 (2013)

\bibitem{hariharan2012discriminative}
Hariharan, Bharath and Malik, Jitendra and Ramanan, Deva:
Discriminative decorrelation for clustering and classification.
ECCV (2012) 459--472


\bibitem{he2015deep}
He, Kaiming and Zhang, Xiangyu and Ren, Shaoqing and Sun, Jian:
Deep Residual Learning for Image Recognition.
arXiv preprint arXiv:1512.03385 (2015)


\bibitem{LSDA}
Judy Hoffman, Sergio Guadarrama, Eric Tzeng, Jeff Donahue, Ross B. Girshick, Trevor Darrell, Kate Saenko:
LSDA: Large Scale Detection Through Adaptation
CoRR abs/1407.5035 (2014)

\bibitem{vgg}
Karen Simonyan and Andrew Zisserman:
Very Deep Convolutional Networks for Large-Scale Image Recognition.
CoRR abs/1409.1556 
(2014)

\bibitem{alexnet}
Krizhevsky, Alex and Sutskever, Ilya and Hinton, Geoffrey E:
Imagenet classification with deep convolutional neural networks.
Advances in neural information processing systems 
(2012) 1097--1105


\bibitem{li2010optimol}
Li, Li-Jia and Fei-Fei, Li:
Optimol: automatic online picture collection via incremental model learning.
International journal of computer vision
\textbf{88} (2010) 147--168

\bibitem{liebelt2010multi}
Liebelt, Joerg and Schmid, Cordelia:
Multi-view object class detection with a 3d geometric model.
Computer Vision and Pattern Recognition (CVPR), 2010 IEEE Conference on
(2010)  1688--1695

\bibitem{lin2015bilinear}
Lin, Tsung-Yu and RoyChowdhury, Aruni and Maji, Subhransu:
Bilinear CNN models for fine-grained visual recognition.
Proceedings of the IEEE International Conference on Computer Vision
(2015) 1449--1457

\bibitem{peng2015learning}
Peng, Xingchao and Sun, Baochen and Ali, Karim and Saenko, Kate:
Learning Deep Object Detectors from 3D Models.
Proceedings of the IEEE International Conference on Computer Vision
(2015) 1278--1286

\bibitem{saenko2010adapting}
Saenko, Kate and Kulis, Brian and Fritz, Mario and Darrell, Trevor:
Adapting visual category models to new domains.
Computer Vision--ECCV (2010) 213--226

\bibitem{schroff2011harvesting}
Schroff, Florian and Criminisi, Antonio and Zisserman, Andrew:
Harvesting image databases from the web.
Pattern Analysis and Machine Intelligence, IEEE Transactions on
\textbf{33} (2011) 754--766

\bibitem{renNIPS15fasterrcnn}
Shaoqing Ren and Kaiming He and Ross Girshick and Jian Sun:
Faster R-CNN: Towards Real-Time Object Detection with Region Proposal Networks.
Advances in Neural Information Processing Systems (NIPS 2015)

\bibitem{simonyan2014two}
Simonyan, Karen and Zisserman, Andrew:
Two-stream convolutional networks for action recognition in videos.
Advances in Neural Information Processing Systems
(2014) 568--576

\bibitem{siva2011weakly}
Siva, Parthipan and Xiang, Tao:
Weakly supervised object detector learning with model drift detection.
IEEE International Conference on Computer Vision (ICCV 2011) 343-350

\bibitem{song2014learning}
Song, Hyun Oh and Girshick, Ross and Jegelka, Stefanie and Mairal, Julien and Harchaoui, Zaid and Darrell, Trevor:
On learning to localize objects with minimal supervision.
arXiv preprint arXiv:1403.1024 (2014)


\bibitem{stark2010back}
Stark, Michael and Goesele, Michael and Schiele, Bernt:
Back to the Future: Learning Shape Models from 3D CAD Data.
BMVC
\textbf{2} (2010)

\bibitem{su2015render}
Su, Hao and Qi, Charles R and Li, Yangyan and Guibas, Leonidas J:
Render for cnn: Viewpoint estimation in images using cnns trained with rendered 3d model views.
Proceedings of the IEEE International Conference on Computer Vision
(2015) 2686--2694


\bibitem{BMVC}
Sun, Baochen and Saenko, Kate:
From Virtual to Reality: Fast Adaptation of Virtual Object Detectors to Real Domains.
BMVC (2014)

\bibitem{sun2014virtual}
Sun, Baochen and Saenko, Kate:
From Virtual to Reality: Fast Adaptation of Virtual Object Detectors to Real Domains.
BMVC (2014)


\bibitem{sun2009multi}
Sun, Min and Su, Hao and Savarese, Silvio and Fei-Fei, Li:
A multi-view probabilistic model for 3d object classes.
Computer Vision and Pattern Recognition, 2009. CVPR 2009. IEEE Conference on
(2009) 1247--1254


\bibitem{GoogleLenet}
Szegedy, Christian and Liu, Wei and Jia, Yangqing and Sermanet, Pierre and Reed, Scott and Anguelov, Dragomir and Erhan, Dumitru and Vanhoucke, Vincent and Rabinovich, Andrew:
Going deeper with convolutions.
arXiv preprint arXiv:1409.4842 (2014)

\bibitem{tenenbaum2000separating}
Tenenbaum, Joshua B and Freeman, William T:
Separating style and content with bilinear models.
Neural computation
(2000) 1247--1283

\bibitem{thomee2015new}
Thomee, Bart and Shamma, David A and Friedland, Gerald and Elizalde, Benjamin and Ni, Karl and Poland, Douglas and Borth, Damian and Li, Li-Jia:
The new data and new challenges in multimedia research.
arXiv preprint arXiv:1503.01817 (2015)

\bibitem{welling2005fisher}
Welling, Max:
Fisher linear discriminant analysis.
Department of Computer Science, University of Toronto
\textbf{3} (2005)

\bibitem{xia2014well}
Xia, Yan and Cao, Xudong and Wen, Fang and Sun, Jian:
Well Begun Is Half Done: Generating High-Quality Seeds for Automatic Image Dataset Construction from Web.
ECCV (2014) 

\bibitem{viscnn}
Zeiler, Matthew D and Fergus, Rob:
Visualizing and understanding convolutional networks.
Computer Vision--ECCV (2014) 818--833

\bibitem{zeiler2014visualizing}
Zeiler, Matthew D and Fergus, Rob:
Visualizing and understanding convolutional networks.
ECCV (2014) 818--833


\bibitem{zhou2015conceptlearner}
Zhou, Bolei and Jagadeesh, Vignesh and Piramuthu, Robinson:
Conceptlearner: Discovering visual concepts from weakly labeled image collections.
Proceedings of the IEEE Conference on Computer Vision and Pattern Recognition
(2015) 1492--1500

\bibitem{zitnick2014edge}
Zitnick, C Lawrence and Doll{\'a}r, Piotr:
Edge boxes: Locating object proposals from edges.
ECCV (2014) 391--405





\end{thebibliography}



\end{document}